\documentclass[journal,10pt]{IEEEtran}

\usepackage{graphicx}
\usepackage{amsmath}
\usepackage{amsfonts}
\usepackage{amssymb}
\usepackage{cite}
\usepackage{setspace}
\usepackage{booktabs}
\usepackage{bm}
\usepackage{array}

\begin{document}

\newcommand{\msub}[2]{#1_{\text{\tiny #2}}}  % macro to put tiny text subscript in math mode

\title{Active Object Localization in Visual Situations}
\author{Max~H.~Quinn, Anthony~D.~Rhodes, and Melanie~Mitchell
\thanks{M. Quinn is with the Department of Computer Science at Portland State University.}
\thanks{A. Rhodes is with the Department of Mathematics and Statistics at Portland State University.}
\thanks{M. Mitchell is with the Department of Computer Science at Portland State University and the Santa Fe Institute.}}

\maketitle

%%%%%%%%%%%%%%%%%%%%%%%%%%%%%%%%%%%%%%%%%%%%%%%%%%%%%%%%%%%%%%%%%%%%%%%%%%%%%%%%%%%%%%%%%%%%%%%%%%%
\begin{abstract}
We describe a method for performing active localization of
objects in instances of visual situations.  A visual situation is an
abstract concept---e.g., ``a boxing match'', ``a birthday party'', ``walking the
dog'', ``waiting for a bus''---whose image instantiations are linked more by
their common spatial and semantic structure than by low-level
visual similarity.  Our system combines given and learned knowledge of
the structure of a particular situation, and adapts that knowledge to
a new situation instance as it actively searches for objects.  More
specifically, the system learns a set of
probability distributions describing spatial and other relationships
among relevant objects. The system uses those
distributions to iteratively sample object proposals on a test image, but also
continually uses information from those object proposals to adaptively
modify the distributions based on what the system has detected.  We test
our approach's ability to efficiently localize objects, using a
situation-specific image dataset created by our group.  We compare the
results with several baselines and variations on our method, and
demonstrate the strong benefit of using situation knowledge and active context-driven
localization.  Finally, we contrast our method with several other
approaches that use context as well as active search for object
localization in images.

\end{abstract}
%%%%%%%%%%%%%%%%%%%%%%%%%%%%%%%%%%%%%%%%%%%%%%%%%%%%%%%%%%%%%%%%%%%%%%%%%%%%%%%%%%%%%%%%%%%%%%%%%%%


\section{Introduction \label{Intro}}

\begin{figure*}[t]
\centering
\parbox{4.2in} {
\includegraphics[width=1in]{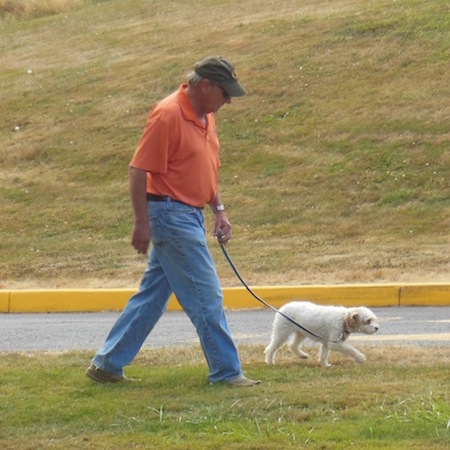}
\includegraphics[width=1in]{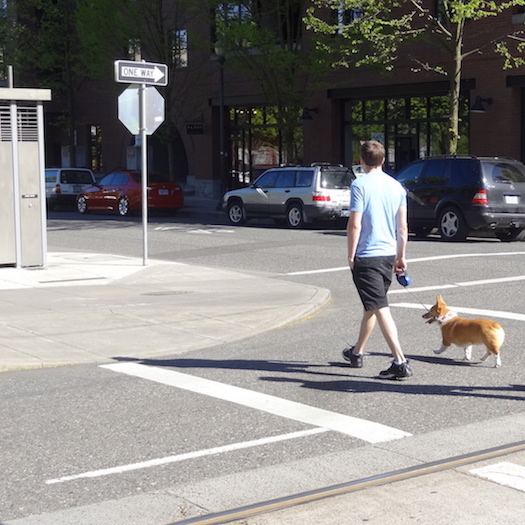} 
\includegraphics[width=1in]{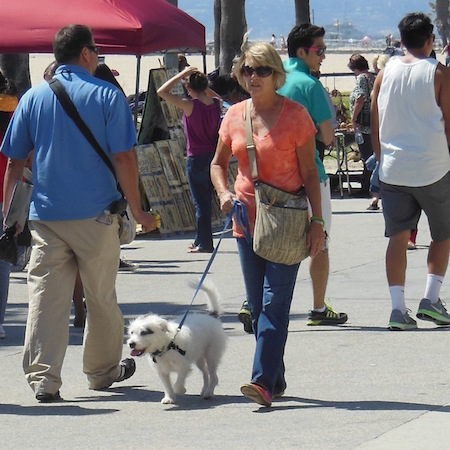}
\includegraphics[width=1in]{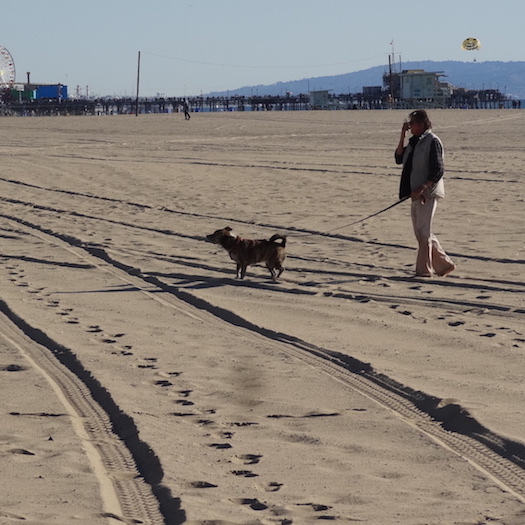}
\caption{Four instances of the ``Dog-Walking'' Situation.  Images are
  from  \cite{PortlandStateDWImages}.  (All figures in this paper are best
  viewed in color.)}
\label{ProtoDW}
}
\quad
\begin{minipage}{2.2in}
\includegraphics[height=1.8in]{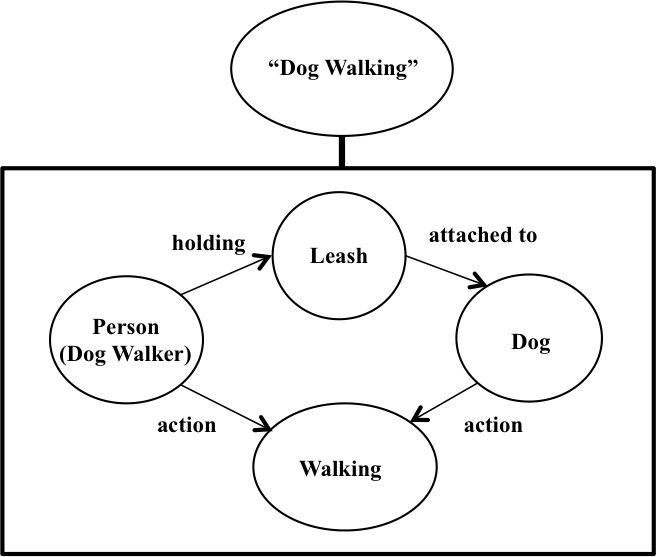} 
\caption{A simple prototype for the Dog-Walking situation.}
\label{Ontology}
\end{minipage}
\end{figure*}

\IEEEPARstart{C}{onsider} the images in Fig.~\ref{ProtoDW}.
Humans with knowledge of the concept ``Walking the Dog'' can easily
recognize these images as instances of that visual situation.  What objects and relationships constitute
this general situation?   A simplified description of the prototypical
``Dog-Walking'' situation might consist of a human dog-walker
holding a leash that is attached to a dog, both of them walking
(Fig.~\ref{Ontology}).  This concept prototype can be mapped
straightforwardly onto the instances in Fig.~\ref{ProtoDW}.

\begin{figure*}[t]
\centering
\includegraphics[width=1in]{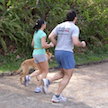}
\includegraphics[width=1in]{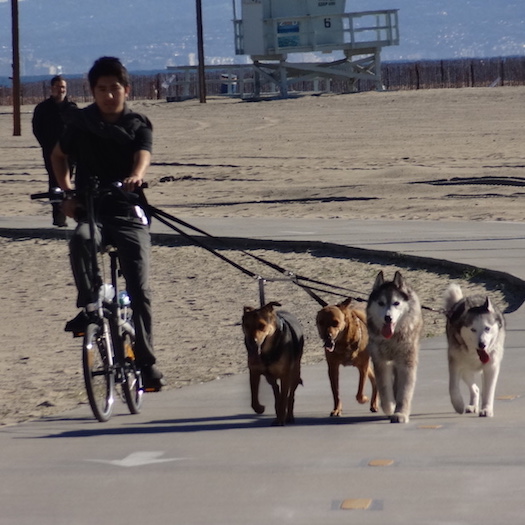}
\includegraphics[width=1in]{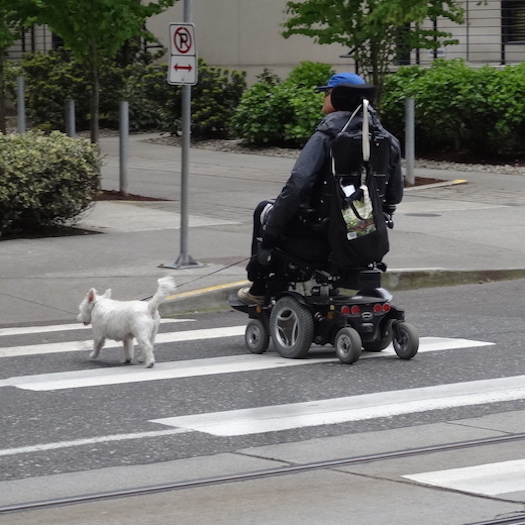} 
\includegraphics[width=1in]{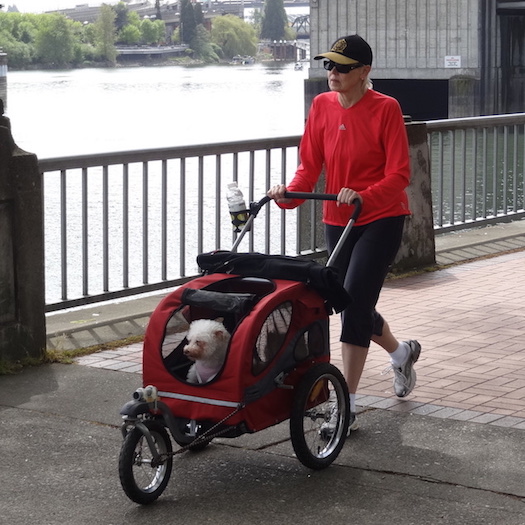}
\includegraphics[width=1in]{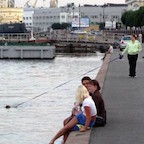}
\includegraphics[width=1in]{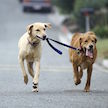}
\caption{Six ``non-prototypical'' \emph{Dog-Walking} situations.  (Images 1--4: \cite{PortlandStateDWImages}.  Image 5: \cite{DogSwimmingImage}.  Image 6: \cite{DogWalkingDogImage}.)} 
\label{NonProtoDW}
\end{figure*}

In general, in order to recognize and understand a particular abstract
visual situation across varied instances, a perceiver must have prior
knowledge of the situation's expected visual and semantic structure,
and must be able to {\it flexibly} adapt that knowledge to a given
input.  Fig.~\ref{NonProtoDW} shows several images semantically
similar to those in Fig.~\ref{ProtoDW}, but which depart in some way
from the prototype of Fig.~\ref{Ontology}.  In some instances there
are not one, but multiple dogs or dog-walkers; in some the dog-``walkers'' and dogs
are not walking, but running (or biking, or swimming, or otherwise riding wheeled
vehicles); in some, the leash is missing.  In the last, the dog-walker
is not a person, but rather another dog.

In spite of these variations, most people would consider all these
(sometimes humorous) images to be instances of the same abstract
category, roughly labeled {\it Dog-Walking}.  This is because people
have the ability to quickly focus on relevant objects, actions,
groupings, and relationships in the image, ignore irrelevant
``clutter,'' and allow some concepts from the prototype to be missing,
or replaced as needed by related concepts.  For example, ``leash'' can
be missing; ``dog'' can be replaced by ``dog-group'' ; ``walking'' can
change to to ``running,'' ``swimming,'' or ``riding''; ``dog-walker
(human)'' can become ``dog-walker (dog),'' and so on.  In
\cite{Hofstadter1995}, Hofstadter et al.\ defined ``conceptual
slippage'' as the act of modifying concepts so as to flexibly map a
perceived situation to a known prototypical situation.  Making
appropriate conceptual slippages is at the core of analogy-making,
which is central to the recognition of abstract
concepts.\cite{Hofstadter2001}

%dog-walking-dog picture: http://www.delcopetcare.com/wp-content/uploads/2013/02/dog-walking.jpg
%dog-swimming-picture: http://www.drollnation.com/gallery/2015/12/randomness-120315-5.jpg

In general, a {\it visual situation} defines a space of visual
instances (e.g., images) which are linked by an abstract concept
rather than any particular low-level visual similarity.  Two instances
of {\it Dog Walking} can be visually dissimilar, but
conceptually analogous.  While the notion of {\it situation}
can applied to any abstract concept \cite{Hofstadter2013}, most people
would consider a visual situation category to be---like {\it
  Dog-Walking}---a named concept that invokes a collection of objects,
regions, attributes, actions, and goals with particular spatial,
temporal, and/or semantic relationships to one another.

The potential open-ended variation in components and relationships
makes it difficult to model abstract situations, such as {\it
  Dog-Walking}, solely by learning statistics from a large collection
of data.  We believe that the process of analogy-making, as developed
in \cite{Hofstadter1995}, is a promising, though yet largely
unexplored method for integrating prior conceptual-level knowledge
with statistical learning in order to capture human-level flexibility
in recognizing visual situations.

We are in the early stages of building a program, called
\emph{Situate}, whose goal is to flexibly categorize and explain new
instances of known visual situations via analogy.  In the long term,
our system will integrate object-recognition and
probabilistic-modeling methods with Hofstadter and Mitchell's \emph{Copycat}
architecture for analogy-making \cite{Hofstadter1994,Mitchell1993}.  Copycat's perceptual process interleaved top-down,
expectation-driven actions with bottom-up exploration in order to
recognize and make analogies between idealized situations.  In
Situate, we are building on Copycat's architecture to apply these
ideas to visual perception and analogy-making.

In our envisioned system, Situate will attempt to make sense of a
given visual image by interpreting it as an instance of a known
situation prototype.  To do this, Situate will locate objects,
attributes, actions, and relationships relevant to the situation and,
when possible, map these, with appropriate slippages, to the situation
prototype.  This mapping will allow Situate to (1) decide if the given
image is an instance of the situation; and (2) if so, to
\emph{explain} how the system interpreted the image as such an
instance.  The explanation will explicitly indicate all the components
of the mapping (including conceptual slippages, if needed) from the
given image to the known prototype.

The problem of recognizing---and explaining---instances of a known
visual situation shares motivation but contrasts with the widely
studied tasks of ``action recognition,'' ``event recognition,'' or
general ``situation recognition'' that have been explored in the
computer vision literature (e.g.,
\cite{Guo2014,Gupta2015,Li2007,Wang2015,Yatskar2016}).  In such tasks,
a system is asked to classify a new image as one of a set of possible
action, event, or situation categories.  While there has been
significant recent progress on particular benchmark datasets, it is
not surprising that such tasks remain very difficult for computer
vision systems, which still perform well below humans.  Moreover, even
when such systems are successful in classifying images as instances of
particular actions or events, it is not clear what such systems
actually ``understand'' about those kinds of situations (though as a
step in this direction, \cite{Gupta2015,Yatskar2016} propose methods
for assigning ``semantic roles'' in instances of action categories).
As Figs.~\ref{ProtoDW} and \ref{NonProtoDW} illustrate, even the
seemingly simple situation of ``Dog-Walking'' turns out to be complex
and open-ended, requiring substantial background knowledge and the
ability to deal with abstract concepts.

Thus, rather than attempting to classify scenes into one of many
situation classes, our current work on Situate focuses on exhibiting
deep knowledge of a {\it particular} situation concept.  Our goal is a
system that can, via analogy, fluidly map its knowledge of a
particular abstract visual concept to a wide variety of novel
instances, while being able to both explain its mapping and to measure
how ``stretched'' that mapping is.  We believe that this approach is
the most likely one to capture general situation-recognition
abilities.  Moreover, we believe that the ability for such
analogy-based explanations may also allow for new methods of training
machine learning systems, such as indicating to a system not only that
a classification was incorrect, but {\it why} it was incorrect.

Developing such a system is a long-term goal.  In this paper we focus
on a shorter-term subtask for Situate: using knowledge of
situation structure in order to quickly locate objects relevant to a
known situation.  We hypothesize that using prior knowledge of a
situation's expected structure, as well as situation-relevant context
as it is dynamically perceived, will allow the system to be accurate
and efficient at localizing relevant objects, even when training data
is sparse, or when object localization is otherwise difficult due to
image clutter or small, blurred, or occluded objects.

In the next section we describe this subtask in more detail.  The
subsequent sections give the details of our dataset and methods,
results and discussion of experiments, an overview of related
literature, and plans for future work.

%%%%%%%%%%%%%%%%%%%%%%%%%%%%%%%%%%%%%%%%%%%%%%%%%%%%%%%%%%%%%%%%%%%%%%%%%%%%%%%%%%%%%%%%%%%%%%%%%%%
\section{Situation Structure and Efficient Object Localization \label{SitStruct}}

For humans, recognizing a visual situation is an active process that
unfolds over time, in which prior knowledge interacts with visual
information as it is perceived, in order to guide subsequent eye
movements.  This interaction enables a human viewer to very quickly
locate relevant aspects of the situation
\cite{Bar2004,Malcolm2014,Neider2006,Potter1975,Summerfield2009}.

Most object-localization algorithms in computer vision do
not exploit such prior knowledge or dynamic perception of
context.  The current state-of-the-art methods employ feedforward deep
networks that produce and test a fixed number of {\it object
  proposals} (also called {\it region proposals})---on the order of
hundreds to thousands---in a given image (e.g.,
\cite{He2015,Girshick2015,Redmon2015}).  An {\it object proposal} is a
region or bounding box in the image (sometimes associated with a
particular object class).  Assuming an object proposal defines a
bounding box, the proposal is said to be a successful localization (or
detection) if the bounding box sufficiently overlaps a target
object's ground-truth bounding box.  Overlap is measured via the
intersection over union (IOU) of the two bounding boxes, and the
threshold for successful localization is typically set to 0.5
\cite{Everingham2010}.

Popular benchmark datasets for object-localization (or
object-detection, which we will use synonymously, although they are
sometimes defined distinctly) include Pascal VOC \cite{Everingham2010}
and ILSVRC \cite{Russakovsky2015}.  In each, the detection task is the
following: given a test image, specify the bounding box of each object
in the image that is an instance of one of $M$ possible object
categories.  In the most recent Pascal VOC detection task, the number
of object categories $M$ is 20; in ILSVRC, $M$ is 200.  For both
tasks, algorithms are typically rated on their {\it mean average
  precision} (mAP) on the task: the {\it average precision} for a
given object category is the area under its precision-recall curve,
and the mean of these values is taken over the $M$ object categories.
On both Pascal VOC and ILSVRC, the best algorithms to date have mAP in
the range of about 0.5 to 0.70; in practice this means that they are
quite good at locating some kinds of objects, and very poor at others.

In fact, state-of-the-art methods are still susceptible to several
problems, including difficulty with cluttered images, small or
obscured objects, and inevitable false positives resulting from large
numbers of object-proposal classifications.  Moreover, such methods require
large training sets for learning, and potential scaling issues as the
number of possible categories increases.

For these reasons, several groups have pursued the more human-like
approach of ``active object localization,'' in which a search for objects
 unfolds over time, with each subsequent time step using information
gained in previous time steps (e.g., \cite{deCroon2011,Gonzalez-Garcia2015,Lu2015}).  

Our approach is an example of active object localization, but in the
context of specific situation recognition.  Thus, only objects
specifically relevant to the given situation are required to be
located.  Situate\footnote{We refer to the system described in this
  paper as ``Situate,'' though what we describe here is only one part
  of the envisioned Situate architecture.} is provided some prior
knowledge---the set of the relevant object categories---and it learns
(from training data) a representation of the expected spatial and semantic structure of the
situation.  This representation consists of a set of joint probability
distributions linking aspects of the relevant objects.  Then, when
searching for the target objects in a new test image, the system
samples object proposals from these distributions, conditioned on what
has been discovered in previous steps.  That is, during a search for
relevant objects, evidence gathered during the search continually
serves as context that influences the future direction of the
search.

Our hypothesis is that this localization method, by combining prior knowledge with
learned situation structure and active context-directed search, will
require dramatically fewer object proposals than methods that do not
use such information.  

In the next sections we describe the dataset we used and
experiments we performed to test this hypothesis.

%%%%%%%%%%%%%%%%%%%%%%%%%%%%%%%%%%%%%%%%%%%%%%%%%%%%%%%%%%%%%%%%%%%%%%%%%%%%%%%%%%%%%%%%%%%%%%%%%%%
\section{Domain and Dataset}

Our initial investigations have focused on the {\it Dog-Walking}
situation category: not only is it easy to find instances to
photograph, but, as illustrated by Figs.~\ref{ProtoDW} and
\ref{NonProtoDW}, it also presents many interesting challenges for the
general tasks of recognizing, explaining, and making analogies between
visual situations.  We believe that the methods developed in Situate
are general and will extend to other situation categories. 

We test our system on a new image dataset, called the ``Portland State
Dog-Walking Images'' \cite{PortlandStateDWImages},  created by our group. 
This dataset currently contains 700 photographs, taken in
different locations by members of our research group. Each image
is an instance of a ``Dog-Walking'' situation in a natural
setting. (Figs.~\ref{ProtoDW} and \ref{NonProtoDW} give some
examples from this dataset.)  Our group has also hand-labeled the
Dog-Walker(s), Dog(s), and Leash(es) in each photograph with tight
bounding boxes and object category labels.\footnote{The photographs
  and label files can be downloaded at the URL given in \cite{PortlandStateDWImages}.  Note that our collection
  is similar to the Stanford 40 Actions \cite{Yao2011}
  ``Walking the Dog'' category, but the photographs in our set are
  more numerous, varied, and have bounding box labels for each
  relevant object.}.

For the purposes of this paper, we focus on a simplified subset of the
{\it Dog-Walking} situation: photographs in which there is exactly one
(human) dog-walker, one dog, one leash, and unlabeled ``clutter''
(such as non-dog-walking people, buildings, etc) as in
Fig.~\ref{ProtoDW}. There are 500 such images in this subset.

%%%%%%%%%%%%%%%%%%%%%%%%%%%%%%%%%%%%%%%%%%%%%%%%%%%%%%%%%%%%%%%%%%%%%%%%%%%%%%%%%%%%%%%%%%%%%%%%%%%
\section{Overview of Our Method \label{OverviewSection}}

Our system's task is to locate three objects---{\it Dog-Walker}, {\it Dog},
and {\it Leash}---in a test image using as few object proposals as possible.
Here, an ``object proposal'' comprises an object category (e.g.,
{\it Dog}), coordinates of a bounding box center, and the bounding box's
size and aspect ratio.

We use the standard criterion for successfully locating an object: an
object proposal's intersection over union (IOU) with the target
object's ground-truth bounding box must be greater than or equal to
0.5.  Our main performance metric is the median number of
object-proposal evaluations per image needed in order to locate all
three objects.

In this section we give an overview of how our system works.  In the
following sections we describe the details of what the system learns,
how it uses its knowledge to search for objects, and the results of
experiments that demonstrate the effect of situation knowledge on the
system's efficiency.

\subsection{Knowledge Given to Situate}
For this task, the only knowledge given to Situate is the set of
relevant object categories $\{${\it Dog-Walker}, {\it Dog}, and {\it
  Leash}$\}$, and the assumption that each image contains exactly one
instance of each of those categories.  In future work we will
investigate less restricted situations, as well as the integration of
richer information about situations, such as additional properties of
constituent objects and more varied relationships between objects.

\subsection{Knowledge Learned by Situate}
From training data (photographs like those in Fig.~\ref{ProtoDW}), our system learns the following probability models:  
\begin{enumerate}
\item {\bf Bounding-Box Size and Shape Priors.} For each of the three
  object categories, Situate learns distributions over bounding-box
  size, which is given as the {\it area ratio} (box area / image area), 
  and shape, which is given as the {\it aspect ratio} (box width / box
  height).  These prior distributions are all independent of one another, and will be used to sample bounding boxes
  in a test image before any objects in that image have been localized.
  (Note that our system does not learn prior distributions over
  bounding-box {\it location}, since we do not want the system to
  model photographers' biases to put relevant objects near the
  center of the image.)

\item {\bf Situation Model.} Situate's learned model of the
  {\it Dog-Walking} situation consists of a set of probability
  distributions modeling the joint locations of the three relevant
  objects, as well as the joint area-ratios and aspect-ratios
  of bounding boxes for the three objects.  These
  distributions capture the expected relationships among the three
  objects with respect to location and size/shape of bounding boxes.
  When an object proposal is labeled a ``detection'' by the system,
  the situation model distributions are re-conditioned on all
  currently detected proposals in order to constrain the expected
  location, size, and shape of the other as-yet undetected objects.

\end{enumerate}
Details of these two learning steps are given in Sec.~\ref{TrainingDetails}. 

\subsection{Running Situate on a Test Image \label{RunningSituate}}

Following the Copycat architecture\cite{Hofstadter1994}, Situate's main data structure is the
{\it Workspace}, which contains the input test image as well as any
object proposals that have scored highly enough.  Situate uses its
learned prior distributions and its conditioned situation model to
select and score object proposals in the Workspace, one at a time.

\begin{figure*}[t]
\centering
\includegraphics{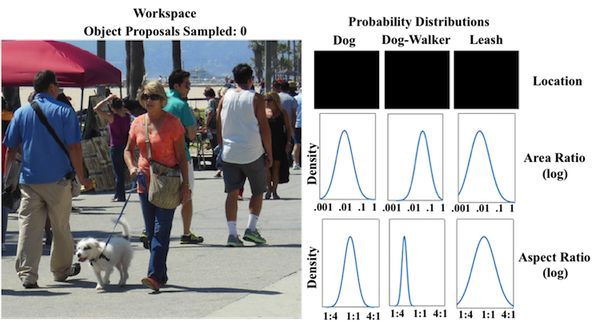}
\caption{Situate's initialization with a test image, before the start
  of a run.  The Workspace is shown on the left, initialized with a
  test image.  On the right, the initial per-category probability
  distributions for location, area ratio, and aspect ratio are shown.
  The (two-dimensional) location distributions for each object
  category are initialized as uniform (here, all black), and the
  area-ratio and aspect-ratio distributions are initialized to the
  learned prior distributions. In this visualization, the density axis has arbitrary units to
  fit the plot size; the intention here is to show the general shape
  of these distributions.  For example, as expected, human dog-walkers
  tend to have higher area ratios and lower aspect ratios than dogs,
  and leashes have considerably more variation than either humans or
  dogs.}
\label{ProbDistInitial}
\end{figure*}

At any time during a run on a test image, each relevant object
category ({\it Dog-Walker}, {\it Dog}, and {\it Leash}) is associated with a
particular probability distribution over locations in the image, as
well as probability distributions over aspect ratios and area ratios for
bounding boxes.  Situate initializes the location distribution for
each object category to uniform over the image, and initializes each object category's
area-ratio and aspect-ratio distribution to the learned prior
distributions (Fig.~\ref{ProbDistInitial}).

At each time step in a run, Situate randomly chooses an object
category that has not yet been localized, and samples from that
category's current probability distributions over location, aspect
ratio, and area ratio in order to produce a new object proposal.  (If
part of the object proposal's bounding box lies outside of the image
boundaries, the bounding box is cropped to the part that lies inside
the image boundaries.)  The resulting proposal is then given a score
for that object category, as described in Sec.~\ref{ScoringSection}.  

The system has two user-defined thresholds: a {\it provisional
  detection threshold} and a {\it final detection threshold}.  These
thresholds are used to determine which object proposals are promising
enough to be added to the Workspace, and thus influence the system's
subsequent search.

If an object proposal's score is greater than or equal to
the {\it final detection threshold}, the system marks the object
proposal as ``final,'' adds the proposal to the Workspace, and
stops searching for that object category.

Otherwise, if an object proposal's score is greater than or equal to
the {\it provisional detection threshold}, it is marked as 
``provisional.''  If its score is greater than any provisional proposal
for this object category already in the Workspace, it replaces that
earlier proposal in the Workspace.  The system will continue searching
for better proposals for this object category.

Whenever the Workspace is updated with a new object proposal, the
system modifies the current situation model to be conditioned on all
of the object proposals currently in the Workspace.  

The conditioned location distributions reflect where the system should
{\it expect} to see instances of the associated object categories,
given the detections so far.  Similarly, the
conditioned area-ratio and aspect-ratio distributions reflect what size and
shape boxes the system should expect for the associated object
categories, given what has already been detected.

The purpose of {\it provisional} detections in our system is to use
information the system has discovered even if the system is not yet
confident that the information is correct or complete.  In
Sec.~\ref{ResultsSection} we describe the results of an experiment that
tests the usefulness of  provisional detections for directing
the system's search for its target objects.

\subsection{Scoring Object Proposals \label{ScoringSection}}

In the experiments reported here, during a run of Situate, each
object proposal is scored by an ``oracle'' that returns the
intersection over union (IOU) of the object proposal with the
ground-truth bounding box for the target object.  The {\it
  provisional} and {\it final} detection thresholds are applied to
this score to determine if the proposal should be added to the
Workspace.  For the experiments described in this paper, we used a
provisional detection threshold of 0.25 and a final detection
threshold of 0.5. This oracle can be thought of as an idealized
``classifier'' whose scores reflect the amount of partial localization
of a target object.

Why do we use this idealized oracle rather than an actual object
classifier?  The goal of this paper is not to determine the quality of
any particular object classifier, but to assess the benefit
of using prior situation knowledge and active context-directed search
on the efficiency of locating relevant objects.  Thus, in this study,
we do not use trained object classifiers to score object proposals.

In future work we will experiment with object classifiers that can
predict not only on the object category of a proposal but also the
amount and type of overlap with ground truth (e.g., see
\cite{Caicedo2015} for interesting work on this topic).

\subsection{Main Loop of Situate \label{MainLoop}}

The following describes the main loop of
Situate, in which the system searches for objects in a new test image.

\begin{list}{}{}
\item {\bf Input:} A test image

\item {\bf Initialization:} Initialize location, area-ratio, and
  aspect-ratio distributions for each relevant object category ({\it
    Dog-Walker}, {\it Dog}, and {\it Leash}).  The initial location
  distributions are uniform; initial area-ratio and aspect-ratio
  distributions are learned from training data.  

\item {\bf Main Loop:  Repeat} for {\it max-num-iterations} or until all three objects are detected: 
\begin{enumerate}
     \item Choose object category $c$ at random from non-detected
       categories in $\{${\it Dog-Walker}, {\it Dog}, and {\it
         Leash}$\}$. 

     \item Sample from category-specific location, area-ratio, and aspect-ratio
       distributions for category $c$ to create an object proposal.

    \item Calculate detection score $d$ for this object proposal.  

    \item If $d \geq \text{{\it final-detection-threshold}}$, mark
      proposal as ``final,'' and add it to the Workspace.  

    \item Else, if
      $d \geq \text{{\it provisional-detection-threshold}}$, mark
      proposal as ``provisional.  If $d$ is greater than the detection
      score of any previous provisional proposal in the Workspace, add
      new proposal to the Workspace, replacing any previous
      provisional proposal for category $c$.  

    \item Update probability distributions in situation model to be conditioned on current proposals in the Workspace.  

 \end{enumerate}

  \item {\bf Return:} number of iterations needed to successfully detect
    all three objects (a {\it completed situation detection}), or, if not successful, ``failure''.  

\end{list}

In our experiments we set {\it max-num-iterations} per image to 1,000. 

%%%%%%%%%%%%%%%%%%%%%%%%%%%%%%%%%%%%%%%%%%%%%%%%%%%%%%%%%%%%%%%%%%%%%%%%%%%%%%%%%%%%%%%%%%%%%%%%%%%
\section{A Sample Run of Situate}

\begin{figure*}[p]
\centering
\includegraphics{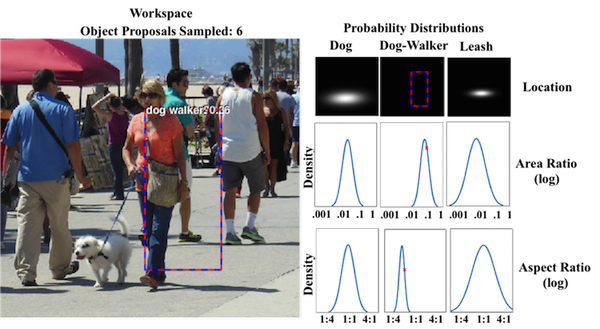}
\caption{The system's state after six object proposals have been
  sampled and scored.  The sixth proposal was for the {\it Dog-Walker}
  category and its score (0.36) is higher than the provisional
  threshold, so a provisional detection was added to the Workspace (red
  dashed box; the samples that gave rise to this proposal are shown in
  red on the various {\it Dog-Walker} probability distributions).  This causes the
  location, area ratio, and aspect ratio distributions for {\it Dog} and
  {\it Leash} to be conditioned on the provisional {\it Dog-Walker} detection,
  based on the learned situation model.  In the location
  distributions, white areas denote higher probability.  The
  area-ratio and aspect-ratio distributions for Dog and Leash have
  also been modified from the initial ones, though the changes are not
  obvious due to our simple visualization.}
\label{ProbDistConditioned1}
\end{figure*}

In this section we illustrate this process with a few visualizations
from a run of Situate---the same run whose initial state is shown in
Fig~\ref{ProbDistInitial}.  Fig.~\ref{ProbDistConditioned1} shows the
state of the system after it has iteratively sampled six object
proposals.  The first five scored below the provisional detection
threshold, but the sixth proposal---for {\it Dog-Walker}---has a score
(0.36) that is above the provisional detection threshold.  Thus a
provisional detection (dashed red box) is added to the Workspace.
This causes the system to modify the current location, area, and
aspect ratio distributions for the other two object categories, so
that they are conditioned on this {\it Dog-Walker} proposal according
to the learned situation model (right side of
Fig~\ref{ProbDistConditioned1}).  The new, conditioned distributions
indicate where, and at what scale, the system should {\it expect} to
locate the dog and the leash, given the (provisionally) detected
dog-walker.  The detected proposal does not affect the distributions
for {\it Dog-Walker}; in our current system, an object category is not
conditioned on itself.

\begin{figure*}[p]
\centering
\includegraphics{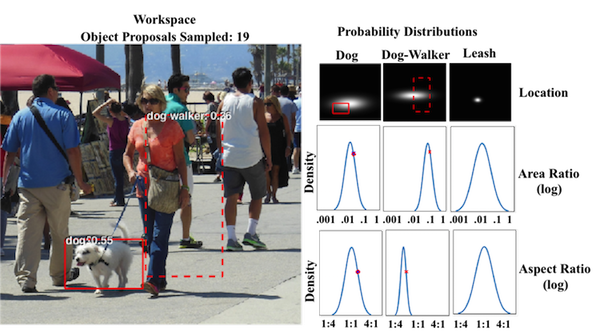}
\caption{The system's state after 19 object proposals have been
  sampled and scored.  The focused conditional distributions have led
  to a {\it Dog} detection at IOU 0.55 (solid red box, indicating
  final detection), which in turn modifies the distributions for {\it
    Dog-Walker} and {\it Leash}.  The situation model is now
  conditioned on the two detections in the Workspace.  Note how
  strongly the {\it Dog} and {\it Dog-Walker} detections constrain the
  {\it Leash} location distributions.}
\label{ProbDistConditioned2}
\end{figure*}

The modified probability distributions for {\it Dog} and {\it Leash}
allow the system to focus its search on more likely locations and box
sizes and shapes.  Fig.~\ref{ProbDistConditioned2} shows the state of
the system after 19 object proposals have been sampled.
The system has successfully located the dog (the IOU with ground truth
is 0.55).  This final detection is added to the Workspace, and the
system will no longer search for dogs.  There are now two detections
in the workspace.  The {\it Leash} probability distributions are
conditioned on both of them, and the {\it Dog} and {\it Dog-Walker}
probability distributions are conditioned on each other (right side of
Fig.~\ref{ProbDistConditioned2}).

\begin{figure*}[p]
\centering
\includegraphics{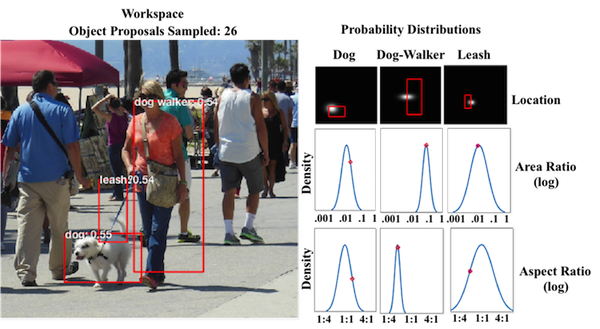}
\caption{The system's state after 26 object proposals have been
  sampled and scored.  Final detections have been made for all three
  objects.}
\label{ProbDistConditioned3}
\end{figure*}

Fig.~\ref{ProbDistConditioned3} shows the system's state after 26
object proposals have been sampled.  The updated situation model
allows a better (and final) {\it Dog-Walker} proposal to be found quickly, as well
as a final {\it Leash} proposal.  

Taken together, Figs.~\ref{ProbDistInitial}--\ref{ProbDistConditioned3} give the
flavor of how Situate is able to use learned situation structure and
active object localization in order to very quickly close in
on the relevant objects in a visual situation.

%%%%%%%%%%%%%%%%%%%%%%%%%%%%%%%%%%%%%%%%%%%%%%%%%%%%%%%%%%%%%%%%%%%%%%%%%%%%%%%%%%%%%%%%%%%%%%%%%%%
\section{Situate's Probabilistic Models: Details \label{TrainingDetails}}

In this section we give details of how our system learns the
probabilistic models of situations it uses to quickly locate relevant objects.

Before learning, all images in the dataset are scaled to have
$\sim$250,000 pixels (preserving each image's original aspect ratio).
The center of the image is assigned coordinates (0,0).  The
coordinates of all ground-truth boxes are converted to the new scale
and coordinate system.

\subsection{Bounding Box Size and Shape Priors \label{BBPriors}}

For each of the three object categories, Situate takes the
ground-truth bounding boxes from the training set, and fits the
natural logarithms of the box sizes (area ratio) and box shapes
(aspect ratio) to normal distributions.  At test time, the system uses
these {\it prior} distributions to sample area ratio and aspect ratio
until one or more detections have been added to the Workspace.

We used log-normal distributions to model these values rather than
normal distributions, because the former are always positive and
give more weight to smaller values.  This made log-normal
distributions a better fit for the data. 

\subsection{Situation Model}

From training data, Situate learns a {\it situation model}: a set of
joint probability distributions that capture the ``situational''
correlations among relevant objects with respect to location, area,
and aspect ratio.  As described above, when running on a test image, Situate will use these
distributions in order to compute category-specific location,
area, and aspect ratio probabilities {\it conditioned} on objects that
have been detected in the Workspace.

The joint probability distributions Situate learns are the following:

\subsubsection*{Location distributions}

Let 
$(\msub{x}{Dog}, \msub{y}{Dog})$, 
$(\msub{x}{DW}, \msub{y}{DW})$, 
$(\msub{x}{L}, \msub{y}{L})$ 
denote the coordinates of the bounding-box center for
a Dog, Dog-Walker, and Leash, respectively.

Situate learns the pairwise and three-way location
relationships among relevant objects, by modeling them as the
following multivariate normal distributions: 

$(\msub{x}{Dog}, \msub{y}{Dog}, \msub{x}{DW}, \msub{y}{DW}) \sim \mathcal{N}(\msub{\bm{\mu}}{Dog,DW},\msub{\bm{\Sigma}}{Dog,DW})$ \\
$(\msub{x}{Dog}, \msub{y}{Dog}, \msub{x}{L}, \msub{y}{L}) \sim \mathcal{N}(\msub{\bm{\mu}}{Dog,L},\msub{\bm{\Sigma}}{Dog,L})$ \\
$(\msub{x}{DW}, \msub{y}{DW}, \msub{x}{L}, \msub{y}{L}) \sim \mathcal{N}(\msub{\bm{\mu}}{DW,L},\msub{\bm{\Sigma}}{DW,L})$ \\
$(\msub{x}{Dog}, \msub{y}{Dog}, \msub{x}{DW}, \msub{y}{DW}, \msub{x}{L}, \msub{y}{L}) \sim \mathcal{N}(\msub{\bm{\mu}}{Dog,DW,L},\msub{\bm{\Sigma}}{Dog,DW,L}).$ 

Here $\bm{\mu}$ is the multivariate mean over locations and $\bm{\Sigma}$ is the
covariance matrix; these parameterize the distribution.

\subsubsection*{Size and Shape Distributions} 
Let $\msub{\alpha}{Dog}$, $\msub{\alpha}{DW}$, and $\msub{\alpha}{L}$
denote the natural logarithm of the bounding box area-ratio for a Dog,
Dog-Walker, and Leash, respectively.  Similarly, let 
$\msub{\gamma}{Dog}$, $\msub{\gamma}{DW}$, and $\msub{\gamma}{L}$
denote the natural logarithm of the bounding box aspect-ratio for a Dog,
Dog-Walker, and Leash, respectively.  

Situate learns the pairwise and three-way size and shape relationships
among relevant objects by modeling them as the following multivariate
log-normal distributions over area ratio and aspect ratio:

$(\msub{\alpha}{Dog}, \msub{\gamma}{Dog}, \msub{\alpha}{DW}, \msub{\gamma}{DW}) \sim \mathcal{N}(\msub{\bm{\mu}}{Dog,DW},\msub{\bm{\Sigma}}{Dog,DW})$ \\
$(\msub{\alpha}{Dog}, \msub{\gamma}{Dog}, \msub{\alpha}{L}, \msub{\gamma}{L}) \sim \mathcal{N}(\msub{\bm{\mu}}{Dog,L},\msub{\bm{\Sigma}}{Dog,L})$ \\ 
$(\msub{\alpha}{DW}, \msub{\gamma}{DW}, \msub{\alpha}{L}, \msub{\gamma}{L}) \sim \mathcal{N}(\msub{\bm{\mu}}{DW,L},\msub{\bm{\Sigma}}{DW,L})$ \\
$(\msub{\alpha}{Dog}, \msub{\gamma}{Dog}, \msub{\alpha}{DW}, \msub{\gamma}{DW}, \msub{\alpha}{L}, \msub{\gamma}{L}) \sim \mathcal{N}(\msub{\bm{\mu}}{Dog,DW,L},\msub{\bm{\Sigma}}{Dog,DW,L}).$ 

Here $\bm{\mu}$ is the multivariate mean over log-area-ratio and
log-aspect-ratio and $\bm{\Sigma}$ is the covariance matrix.  

We chose normal and log-normal distributions to use in the Situation
Model because they are very fast both to learn during training
and to condition on during a run on a test image.  Their efficiency
would also scale well if we were to add new attributes to the
situation model (e.g., object orientation) and thus increase the
dimensionality of the distribution.

However, such distributions sometimes do not capture the situation
relationships very precisely.  We discuss limitations of these
modeling choices in Sec.~\ref{Conclusions}.

All of the code for this project was written in MATLAB, and will be
released upon publication of this paper.

%%%%%%%%%%%%%%%%%%%%%%%%%%%%%%%%%%%%%%%%%%%%%%%%%%%%%%%%%%%%%%%%%%%%%%%%%%%%%%%%%%%%%%%%%%%%%%%%%%%
\section{Comparison Methods \label{ComparisonMethodsSection}}

In Sec.~\ref{SitStruct}, we stated our hypothesis: by using prior
(given and learned) knowledge of the structure of a situation and by
employing an active context-directed search, Situate will require
dramatically fewer object proposals to locate relevant objects than
methods that do not use such information.  In order to test this
hypothesis, we compare Situate's performance with that of four
baseline methods and two variations on Situate.  These are described in this section.

%%%%%%%%%%%%%%%%%%%%%%%%%%%%%%%%%%%%%%%%%%%%%%%%%%%%%%%%%%%%%%%%%%%%%%%
\subsection{Uniform Sampling \label{UniformSubsection}}
As the simplest baseline, we use uniform sampling of location, area
ratio, and aspect ratio to form object proposals.  Uniform sampling
uses the same main loop of Situate (cf. Sec.~\ref{MainLoop}) but
keeps the location, area ratio, and aspect ratio distributions fixed
throughout the search, as follows.  Given an object category $c \in
\left\{\text{{\it Dog-Walker}, {\it Dog}, {\it Leash}}\right\}$, an object proposal is formed by sampling the following values: 
\begin{itemize}
\item {\it Location}: The coordinates $(x,y)$ of the bounding-box center are sampled uniformly across the entire image, independent of $c$.  

\item {\it Area Ratio}: A value $\alpha$ is sampled uniformly in
  the range $\lbrack\ln .01 , \ln .5\rbrack$; the area ratio is set to
  $e^{\alpha}$.  This means that the area ratio will be between 1\%
  and 50\% of the image size, with higher
  probability given to smaller values.  This gives a reasonable fit to
  the distribution of area ratios (independent of $c$) seen in the training set.

\item {\it Aspect Ratio}: A value $\gamma$ is sampled uniformly
  in the range $\lbrack\ln(.25),\ln(4)\rbrack$; the area ratio is set to
  $e^{\gamma}$. This gives a reasonable fit to the distribution of
  aspect ratios (independent of $c$) seen in the training set.
\end{itemize}
%%%%%%%%%%%%%%%%%%%%%%%%%%%%%%%%%%%%%%%%%%%%%%%%%%%%%%%%%%%%%%%%%%%%%%%
\subsection{Sampling from Learned Area-Ratio and Aspect-Ratio Distributions}

A second baseline method samples location uniformly (as in the uniform
method described above), but samples area ratio and aspect ratio from
the prior (learned) per-category distributions
(cf. Sec.~\ref{BBPriors}), keeping all these distributions fixed
throughout the search.  This method tests the usefulness of learning
prior log-normal distributions of box area and aspect ratio.

%%%%%%%%%%%%%%%%%%%%%%%%%%%%%%%%%%%%%%%%%%%%%%%%%%%%%%%%%%%%%%%%%%%%%%%
\subsection{Location Prior: Salience}

As we mentioned above, our model does not include a learned prior
distribution on location, because we do not want to model the
photographer bias which tends to put relevant objects in the center of
a photo.

To test a baseline method with a location prior, we used a
fast-to-compute salience algorithm, similar to the one proposed in
\cite{Itti1998} (and extended in \cite{Elazary2010}).  In particular,
when given an image, our system creates a category-independent location prior by computing
a salience map on each test image, and then normalizing it to form a
probability distribution over location---the {\it salience prior}.  Each
pixel is thus given a value $p \in \lbrack0,1\rbrack$ representing the
system's assessment of the probability that it is in the center of a
foreground ``object'' (independent of object category).

This method uses the main loop of Situate, but throughout the search
the location distributions are fixed to this salience prior; the
bounding box size and shape distributions are fixed to the uniform
distributions described in Sec.~\ref{UniformSubsection}.

In more detail, following \cite{Itti1998}, our salience computation
decomposes an image into a set of feature maps, each indicating the
presence of a low-level visual feature at a particular scale, and then
integrates the resulting feature maps into a single salience map. The
low-level features are inspired by features known to be computed in
the primary visual cortex \cite{Itti2001} including local intensity,
color contrast, and presence of edge orientations.  Our method differs from that
of \cite{Itti1998} only in parameter settings that allow us to
decompose an image into feature maps quickly, to avoid aliasing
artifacts and bias, and to avoid unrepresented edge orientations and
spatial frequencies.  We found that much of the salience in the
resulting maps is located on the edges of objects, which is
reasonable.  However, because our purpose is to
select points near the center of mass of objects, we smooth the 
resulting salience map with a Gaussian kernel with
standard deviation approximately 10\% of the image width.  A similar
variation of \cite{Itti1998} was recently shown to compare well with
other state-of-the-art saliency methods \cite{Frintrop2015}.
	
%%%%%%%%%%%%%%%%%%%%%%%%%%%%%%%%%%%%%%%%%%%%%%%%%%%%%%%%%%%%%%%%%%%%%%%%%%%%%%%%%%%%%%%%%%%%%%%%%%%
\subsection{Randomized Prim's Object Proposals} 

As a final baseline for comparison, we use the Randomized Prim's
Algorithm \cite{Manen2013}, a category-independent object-proposal
method.  Given an image, this method first creates a superpixel
segmentation, and then creates object proposals by repeatedly
constructing partial spanning trees of the superpixels. The nodes of
the partial spanning trees are individual superpixels, and the edges
are weights based on superpixel similarity, measured along several
dimensions (e.g., color, common border ratio, and size).  The
algorithm constructs a partial spanning tree by starting with a single
randomly chosen node, and then choosing new nodes to add to the tree
probabilistically based on edge weight.  A randomized stopping
criterion is used to terminate a partial spanning tree.  The final
object proposal bounding-box is constructed to surround the
superpixels in the tree.  This procedure is repeated $N$ times, where the
number $N$ of object proposals requested per image is a user-defined
parameter.  The number of proposals generated is not always equal to
$N$ since near-duplicate proposals are removed from the set of
generated proposals.  Note that the proposals produced are not ranked
by the algorithm in any way.  This algorithm was found to be
competitive with several other 2014 state-of-the-art object proposal
methods on Pascal VOC data \cite{Hosang2014}.

To compare this method with Situate, we used the MATLAB implementation
provided at https://github.com/smanenfr/rp.  We used the value 200 for
the ``minimum number of superpixels,'' and $N=10,000$.  In order to
get a large-enough set of bounding boxes, we ran the algorithm four
times on each image, each time using a different color space (HSV,
LAb, Opponent, and rg) for the color similarity feature (see
\cite{Manen2013} for details).

For each image in the test set, we randomly sampled 1,000 object
proposals, one at a time (without replacement) from the generated set.
If a sampled proposal bounding box had IOU $\geq$ 0.5 for any of the
three relevant ground truth objects, that object was marked as
``final,'' indicating a final detection.  If all three objects were
detected (marked as ``final'') during the sampling procedure, sampling
was stopped and the number of proposals sampled was returned.
Otherwise ``failure'' was returned after 1,000 proposals were sampled.

%%%%%%%%%%%%%%%%%%%%%%%%%%%%%%%%%%%%%%%%%%%%%%%%%%%%%%%%%%%%%%%%%%%%%%%%%%%%%%%%%%%%%%%%%%%%%%%%%%%
\subsection{Combining Salience with Learned Distributions}

In addition to these baselines, we also investigated two variations on
our method.  In the first, we experimented with a combination of
methods, as follows.  The prior location distributions for all
categories are based on salience, and the prior distributions for
bounding box size and shape are based on the learned log-normal
distributions.  After one or more detections are added to the
Workspace, the conditional distributions associated with the situation
model are used as described in previous sections, but with one change:
the conditioned location distribution for each category is combined
with the prior salience distribution by pointwise multiplication,
followed by addition of a small non-zero value to each pixel (to avoid
zero-probability locations), followed by renormalization to produce a
new location distribution.

%%%%%%%%%%%%%%%%%%%%%%%%%%%%%%%%%%%%%%%%%%%%%%%%%%%%%%%%%%%%%%%%%%%%%%%%%%%%%%%%%%%%%%%%%%%%%%%%%%%
\subsection{No Provisional Detections }
In a second variation, we removed the system's ability to use
provisional detections to condition situation model distributions.
This means that the only kind of detections are final detections,
which require $IOU \geq 0.5$.  In this experimental condition, partial
detections ($IOU \leq 0.5$) provide no information to the system.
This condition tests the usefulness of using such partial information.
We ran this using the ``Combined Learned Distributions and Salience''
version of our system, since that method exhibited the best
performance, as we show in the next section.

%%%%%%%%%%%%%%%%%%%%%%%%%%%%%%%%%%%%%%%%%%%%%%%%%%%%%%%%%%%%%%%%%%%%%%%%%%%%%%%%%%%%%%%%%%%%%%%%%%%
\section{Results \label{ResultsSection}}   

In this section we present results from running the methods described
above.  In reporting results, we use the term {\it completed situation
  detection} to refer to a run on an image for which a method
successfully located all three relevant objects within 1,000 iterations; we use the term {\it
  failed situation detection} to refer to a run on an image that did
not result in a completed situation detection within 1,000 iterations.

The various methods described above are characterized by: (1) {\bf
  Location Prior:} whether the prior distribution on location is
uniform or based on salience; (2) {\bf Box Prior:} whether the prior
distributions on bounding-box size and shape are uniform or learned;
and (3) {\bf Situation Model:} whether, once one or more object
detections are added to the Workspace, a learned situation model
conditioned on those detections is used instead of the prior
distributions, and whether that conditioned model is combined with
a salience prior for location. 

As described above, our dataset contains 500 images.  For each method,
we performed 10-fold cross-validation: at each fold, 450 images were
used for training and 50 images for testing.  Each fold used a
different set of 50 test images.  For each
method we ran the algorithm described in Sec.~\ref{RunningSituate} on the test images, with
{\it final-detection-threshold} set to 0.5, {\it
  provisional-detection-threshold} set to 0.25, and {\it maximum
  number of iterations} set to 1,000.  In reporting the results, we
combine results on the 50 test images from each of the 10 folds and
report statistics over the total set of 500 test images.

%%%%%%%%%%%%%%% RESULT1: MEDIAN TIME TO COMPLETE DETECTIONS %%%%%%%%%%%%%%%%%
\subsection{Number of Iterations Per Image to Reach Completed Situation Detection}

\begin{figure*}[t]
\centering
\includegraphics[width=4in]{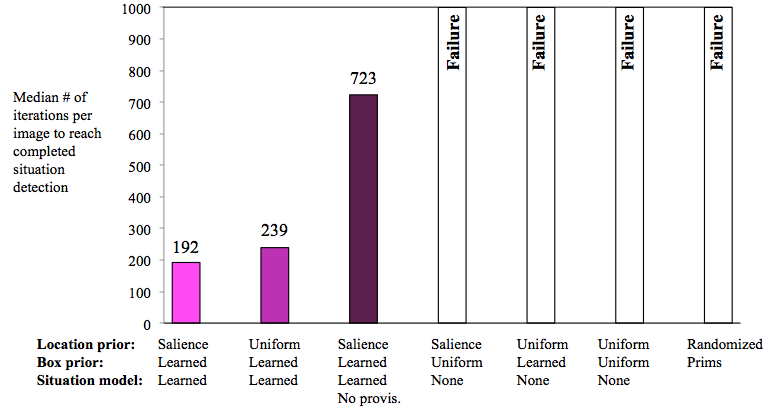}
\caption{Results from seven different methods, giving median number of
  iterations per image to reach a completed situation
  detection (i.e., all three objects are detected at final detected
  threshold).  If a method failed to reach a completed situation
  detection within the maximum iterations on a majority of
  test images, its median is given as ``Failure''.}
\label{Results1}
\end{figure*}

Fig.~\ref{Results1} gives, for each method, the median number of
iterations per image in order to reach a completed situation
detection. The medians are over the union of test images from all 10
folds---that is for 500 images total.  The median value is given as
``failure'' for methods on which a majority of test image runs
resulted in failed situation detections.  We used the median instead
of the mean to allow us to give a statistic that includes the
``failure'' runs.

%%%%%%%%%%%%%%% RESULT2: CUMULATIVE DETECTIONS PLOT %%%%%%%%%%%%%%%%%

\begin{figure*}[t]
\centering
\includegraphics[width=5in]{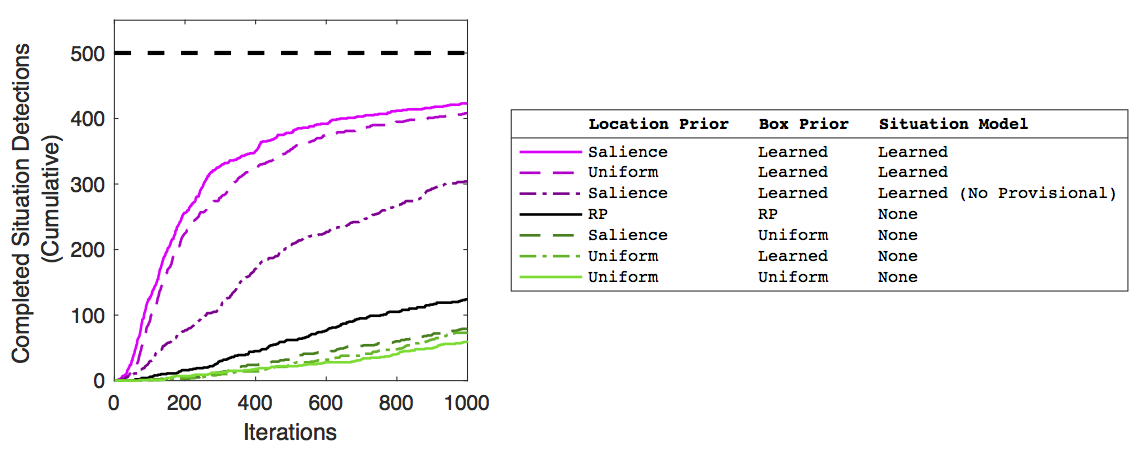}
\caption{Cumulative number of completed situation detections as a
  function of iterations.  For each value $n$ on the $x$-axis, 
  the corresponding $y$-axis value gives the number of test image runs reaching completed situation detections with 
  $n$ or fewer object proposal evaluations.  ``RP'' refers to the Randomized Prim's algorithm.}
\label{Results2}
\end{figure*}

Fig.~\ref{Results1} shows the strong benefit of using a situation
model. The most effective method was the combination of a salience
prior, learned box priors, and a learned situation model.  For a
majority of test images this method required fewer than 200 iterations
to locate all three objects.  The same method without salience
performed only slightly worse, indicating that the salience prior had
only a minor effect on the system's efficiency.  It is also clear that
using provisional detections to guide search is very helpful; removing
this ability (the ``No provis.'' case) nearly quadrupled the median
number of iterations needed.  The four comparison methods with no
situation model failed to reach a completed situation detection on a
majority of the test images.

A more detailed way to visualize results from our runs is given in
Fig.~\ref{Results2}.  This plot shows, for each method, a cumulative
count of completed situation detections as a function of number of iterations.  In particular, it shows, for each number
$n$ of iterations, how many of the 500 test images had complete
situation detections within $n$ or fewer iterations.
For example, for the best-performing method (``Salience, Learned,
Learned''), about 250 test images had complete situation detections within
200 or fewer iterations.  The two top methods---those using
learned situation models that were influenced by provisional
detections---both show a steep rise in the initial part of the curve,
which means that for most of the test images only a small
number of object proposal evaluations were needed.  

Fig.~\ref{Results2} shows that even the best method failed to reach
complete situation detections on about 75 of the 500 test images.  By
looking at the final state of the Workspace on these images, we found
that the most common problem was locating leashes.  In many of the
failure cases, a partial leash was detected, but the system was unable
to improve on the partial detection.  

%%%%%%%%%%%%%%% RESULT3: DETECTION DELTAS %%%%%%%%%%%%%%%%%
\subsection{Sequence of Detections}

The previous plots showed results for completed situation detections.
What can we say about the sequence of object detections within each image?
For methods using a situation model, we would expect that detecting
the first object would take the most iterations. This is because, before a first object is detected, the
system relies only on its prior distributions.  But once a first
object is detected, the situation model, conditioned on that
detection, serves to constrain the search for a second object, and it
should be detected much more quickly than the first object.  Once a
second object is detected, the situation model is conditioned on two
detections, constraining the search even more, so we might expect the
third detection to occur even more quickly.

\begin{figure*}[t!]
\centering
\includegraphics[width=4in]{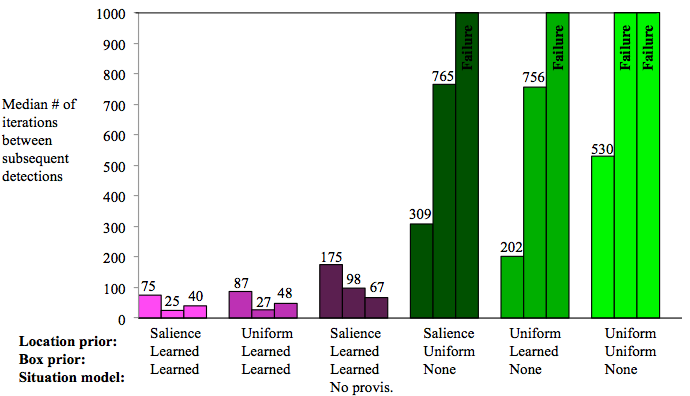}
\caption{Results from different methods giving median number of
  iterations between subsequent object detections (at the {\it
    final-detection threshold}.  For each method, the plot gives three
  bars: the first bar is $\tilde{t}_{0,1}$, the median number of iterations to the first object
  detection in that image; the second bar is $\tilde{t}_{1,2}$, the median number of iterations from the first
  to the second detection; and the third bar is $\tilde{t}_{2,3}$, the median number of iterations from the
  second to the third detection. Results for Randomized Prims is not
  included here because we did not collect this finer-grained data
  for that method.}
\label{Results3}
\end{figure*}

We tested these expectations by recording, for each test image, the
number of iterations to the first detection, from the first to the
second, and from the second to the third.  We denote these as
$t_{0,1}$, $t_{1,2}$, and $t_{2,3}$, respectively.  Here, by ``detection,'' 
we refer to detections at the {\it final-detection-threshold}.  Each method's median
values of $t_{0,1}$, $t_{1,2}$, and $t_{2,3}$ over the 500 test images are shown in
Fig.~\ref{Results3}.  As before, if a method fails to achieve a first,
second, or third detection within 1,000 iterations, that detection is
marked as ``failure''.  In Fig.~\ref{Results3}, a bar marked as
``Failure'' indicates that the method failed on a majority of the test
images.  (Results for Randomized Prims are not included here because
we did not collect this finer-grained data for that method.)

Let $\tilde{t}_{0,1}$ denote the median time to the first detection, $\tilde{t}_{1,2}$
denote the median time from the first to the second detection, and
$\tilde{t}_{2,3}$ denote the median time from the second to the third
detection, for a given method.  For the three methods using a
situation model, Fig.~\ref{Results3} shows that, as expected,
$\tilde{t}_{1,2}$ and $\tilde{t}_{2,3}$ are both considerably shorter than $\tilde{t}_{0,1}$.
However, surprisingly, for the first two methods, $\tilde{t}_{2,3}$ is larger
than $\tilde{t}_{1,2}$.  This appears to be 
because dog-walkers (i.e., humans) are usually the first object to be detected (since they
are the largest of the three objects and also the ones that best fit the learned probability distributions).  
dogs are usually detected second, and leashes usually detected last.  Leashes tend to be the most
difficult objects to detect, because they are in many cases quite small or
otherwise are outliers with respect to the normal distributions
learned from the training data.  Conditioning on a learned situation
model helps a great deal in locating these objects---after
all, $\tilde{t}_{2,3}$ is relatively quite low---but the difficulty of
detecting some leashes is most likely what causes $\tilde{t}_{2,3}$ to be
larger than $\tilde{t}_{1,2}$ for the first two methods. 

In the methods that do not use a situation model, $\tilde{t}_{1,2}$ and
$\tilde{t}_{2,3}$ are much higher than $\tilde{t}_{0,1}$, which reflects the relative
difficulty of detecting the three different object categories without
the help of situational context.

%%%%%%%%%%%%%%% DISCUSSION %%%%%%%%%%%%%%%%%
\subsection{Discussion}

The results presented in this section strongly support
our hypothesis that by using knowledge of a situation's structure in
an active search, our method will require dramatically fewer object
proposals than methods that do not use such information.  Situate's
active search is directed by a set of probability models that are
continually updated based on information gained by the system as it
searches.  Our results show that using information from provisional,
incomplete detections is key to closing in quickly on a complete
situation detection.  

Our results showed that a location prior based on a fast-to-compute
salience map only marginally improved the speed of localization.  More
sophisticated salience methods might reduce the number of iterations
needed, but the computational expense of those methods themselves
might offset the benefits.  This is something we plan to explore in
future work.

Since we used an idealized oracle to score object proposals, the
results we report here support our hypothesis only if we assume that
the system receives feedback about partial detections.  In future work
we will experiment with using a object classifier that is trained to
provide feedback about the proposal's likely overlap with a target
ground-truth region.

In addition, we found that, for the best-performing method, the
failures to reach completed situation detections were largely due to
the difficulty of locating leashes.  This is an example of an object
category that is very hard to locate without context, and even with
strong contextual information it is often hard to go beyond a partial
detection.  To do this, it will be important to identify small object
interactions, such as a person's hand holding a leash.  Recent work in
computer vision has addressed this kind of issue (e.g.,
\cite{Oramas2015,Rosenfeld2016,Singh2016,Yao2010b}) and it will be one
focus of our future research. 

%%%%%%%%%%%%%%%%%%%%%%%%%%%%%%%%%%%%%%%%%%%%%%%%%%%%%%%%%%%%%%%%%%%%%%%%%%%%%%%%%%%%%%%%%%%%%%%%%%% 
\section{Related Work}
In this section we review work that is most closely related to our
goals and methods: using context to localize objects, and active
object localization.

\subsection{Related Work on using Context to Localize Objects}

Until recently the dominant paradigm in object localization and
detection has been the exhaustive ``sliding windows'' approach (e.g.,
\cite{Felzenszwalb2010}).  In the last several years, more
selective approaches to creating category-independent object proposals
have become popular (e.g., \cite{Girshick2014}).  However, in order to make
object detection more efficient and accurate, many groups have looked
at how to use contextual features.

The term ``context'' takes on different meanings in different papers
\cite{Galleguillos2010}.  In some cases it refers simply to
co-occurrence of objects, possibly in the same vicinity.  For example,
\cite{Felzenszwalb2010} uses co-occurrence statistics to re-score
candidate bounding boxes (``sheep'' and ``cow'' are more likely to
co-occur in an image than "sheep" and "bus").  Several groups (e.g.,
\cite{Choi2010,Cao2015,Divvala2009,Li2007,Galleguillos2008}) use graphical models to
capture statistics related to co-occurrence well as some spatial
relationships among objects in an image.

Other approaches use ``Gist'' and other global image features to relate
overall scene statistics or categories to object locations, sometimes
combining these features with co-occurrence statistics and spatial
relationships between objects (e.g.,
\cite{Marat2012,Mottaghi2014,Szegedy2014,Torralba2003}).  Still other
methods learn statistical relationships between objects and features
of pixel regions that surround the objects (e.g.,
\cite{Gould2008,Zhu2015}) or objects (``things'') and homogeneous image
regions (``stuff'') \cite{Heitz2008}.  Going further,
\cite{Izadinia2014} describes a method for learning joint
distributions over scene categories (e.g., ``street,'' ``office,''
``kitchen'') and context-specific appearance and layout of objects in
those scenes.

More recently, several groups have combined object-proposal methods
such as R-CNN \cite{Girshick2014,Girshick2015} with graphical models
(e.g., \cite{Chu2016}) or other CNN architectures (e.g.,
\cite{Gupta2015}) to combine object-to-object context with whole scene
context in order to improve object detection.

The context-based methods described above are only a sample of the
literature on incorporating context into object detection methods, but
they give the general flavor of techniques that have been explored.
Our work shares several goals with these approaches, but differs in at
least four major aspects. One is our focus on images that are
instances of abstract visual situations such as ``Dog-Walking,'' in
which ``situational'' context from relevant objects helps very
significantly in object localization, as was shown in
Sec.~\ref{ResultsSection}.  In contrast, the approaches described above have
been designed and tested using image datasets in which the role of
semantic or ``situational'' context is limited.  This means that
context of the kinds described above most often do not result in large
improvements in localization accuracy \cite{Choi2010}.

A second aspect in which our work differs from other work on context
is that we use prior situation knowledge to focus on specific object
categories relevant to the given situation.  The methods described
above typically collect statistics on contextual relationships among
{\it all} possible object categories (or at least all known to the
system).  The idea is that the system will learn which object
categories have useful contextual relationships.  This is an admirable but quite difficult task, and it entails 
some risks: with many categories and without enough training data,
the system may learn false relationships among unrelated categories,
and as the number of known categories increases, the amount of
training data and computation needed for training and testing may
scale in an untenable way.  Our approach assumes that the system has
some prior knowledge of a situation's relevant concepts; Situate is not
meant to be a general statistical ``situation-learning'' system.

A third point of difference is the importance in our system of
explanation to the user.  As we described in Sec~\ref{Intro}, an
key aspect of our envisioned system is its ability to explain its
reasoning to humans via the structures it builds in the Workspace and
their mappings to (and possible slippages from) a prototype situation.
Creating an explicit situation model is part of the system's ability
to explain itself.  For example, visualizations like those in
Figs.~\ref{ProbDistConditioned1}-\ref{ProbDistConditioned3} make it
very clear how the system's perception of context facilities object
localization.  In contrast, most other recent context-based object
detection papers simply report on the difference context makes in
per-category average precision---it is increased for some categories,
decreased for others, but it is usually not clear why this happens, or
what could be done to improve performance.

Finally, unlike our system, the approaches to context described above
are not dynamic---that is, context is used on a test image as an added
feature or variable used to optimize object detection.  In our system,
contextual features change over time as a result of what the system
perceives; there is temporal feedback between perception of context and
object localization.  This kind of feedback is the hallmark of active
object localization, which leads to our discussion in the next
subsection.

%%%%%%%%%%%%%%%%%%%%%%%%%%%%%%%%%%%%%%%%%%%%%%%%%%%%%%%%%%%%%%%%%%%%%%%%%%%%%%%%%%%%%%%%%%%%%%%%%%%
\subsection{Related Work on Active Object Localization}

Our method is an example of active object localization, an iterative
process inspired by human vision, in which the evaluation of an object
proposal at each time step provides information that the system can
use to create a new object proposal at the next time step.  

Work on active object localization has a long history in computer
vision, often in the context of active perception in robots
\cite{Ballard1991} and modeling visual attention \cite{Borji2013}. The
literature of this field is large and currently growing---here we
summarize a few examples of recent work most similar to our own.

Alexe et al. \cite{Alexe2012} proposed an active, context-driven
localization method: given an initial object proposal (``window'') in
a test image, at each time step the system uses a nearest-neighbor
algorithm to find training image regions that similar in position
and appearance to the current object proposal.  These nearby training
regions then ``vote'' on the next location to propose in the test
image, given each training region's displacement vector relative
to the ground-truth target object.  These ``vote-maps'' are integrated
from all past time steps; the integrated vote map is used to choose
the next object proposal.  The system returns the highest scoring of
all the visited windows, using an object-classifier score.  The
authors found that their method increased mAP on the Pascal VOC 2010
dataset by a small amount over the state of the art method of the
time, but used only about one-fourth the window evaluations as that
method.  However, the nearest-neighbor method can be costly in terms
of efficiency.  A more efficient and accurate version of
this method, using R-CNN object proposals and random forest
classifiers is described in \cite{Gonzalez-Garcia2015}.

Some groups have used recurrent neural networks (RNNs) to perform
active localization.  For example the work of Mnih et
al. \cite{Mnih2014} (extended in \cite{Ba2014}) combines a
feedforward``glimpse network,'' which extracts a representation of an
image region, with an RNN that inputs that representation as well as
its own previous state to produce a new state that is input to an
``action network'' which outputs the next location to attend to.  the
algorithms described in \cite{Mnih2014} and \cite{Ba2014} are tested on cluttered
and translated MNIST digits as well as images of house numbers from
Google Street View.

Several groups frame active object localization as a Markov decision
process (MDP) and use reinforcement learning to learn a search policy.  The
approach proposed in \cite{Caicedo2015} involves learning a search
policy for a target object that consists of a sequence of bounding-box
transformations, such as horizontal and vertical moves, and changes to
scale and aspect ratio.  The algorithm starts with a bounding box
containing most of the image, and uses a deep reinforcement learning
method to learn a policy that maps bounding box features to actions,
obtaining a reward if IOU with the ground truth bounding box is
reduced by the action.  Once learned, this policy is applied to new
test images to locate target objects.  The authors tested this
method on the Pascal VOC 2007 dataset; it obtained competitive mAP
using on average a very small number of policy actions.

In the MDP method proposed in \cite{Chen2016}, an action consists of
running a detector for a ``context class'' that is meant to help
locate instances of the target ``query class''.  To locate appropriate
``context regions'' in a test image, the system relies on stored pairs
of bounding boxes from co-occurring object classes in training images,
along with their displacement vector and change of aspect ratio.  If
the policy directs the system to detect a given context class in a
test image, the system uses a nearest-neighbor method to create a vote
map similar to that proposed in \cite{Alexe2012}, as described above.
This method was tested on the Pascal VOC 2010 dataset and exhibited a
small increase in mAP over other state-of-the-art methods while
requiring significantly fewer object-proposal evaluations

Nagaraja et al. \cite{Nagaraja2015} proposed an MDP method in which a
search policy is learned via ``imitation learning'': in a given state,
an oracle demonstrates the optimal action to take and the policy
subsequently learns to imitate the oracle.  The algorithm starts with
a set of possible object proposals (generated by a separate algorithm)
and its learned policy guides exploration of these proposals.  The
system was tested on a dataset of indoor scenes and was found to
improve average precision (as a function of number of proposal
evaluations) on several object categories as compared to a simple
proposal-ranking method.

Like these methods, our approach focuses on perception as a temporal
process in which information is used as it is gained to narrow the
search for objects.  However, the differences between our approach and
these other active localization methods are similar to those we
described in the previous subsection: often these methods are tested
on datasets in which the role of context is limited; these methods
often rely on exhaustive co-occurrence statistics among object
categories; and it is usually hard to understand why these methods
work well or fail on particular object categories.  In addition, the
reinforcement-learning based methods learn a policy that is fixed at
test time; in our method, the representation of a situation itself
adapts (via modifications to probability distributions) as information
is obtained by the system.  Finally, the amount of training data and
computation time required can be quite high, especially for
reinforcement learning and RNN-based methods.  

In future work, we plan to compare some of these methods with ours on
our situation-specific dataset, in order to assess the performance of
these approaches on images in which context plays a large role, and to
assess the relative requirements for training data and computation
time among the different methods.

%%%%%%%%%%%%%%%%%%%%%%%%%%%%%%%%%%%%%%%%%%%%%%%%%%%%%%%%%%%%%%%%%%%%%%%%%%%%%%%%%%%%%%%%%%%%%%%%%%%

\section{Conclusions and Future Work \label{Conclusions}}

Our work has provided the following contributions: 
\begin{itemize}
\item We have proposed a new approach to actively localizing objects
  in visual situations, based on prior knowledge, adaptable
  probabilistic models, and information from provisional detections.  

\item We created a new situation-specific dataset (the Portland State
  Dog-Walking Images) with human-labeled object bounding boxes, and
  made it publicly available.

\item We performed experiments comparing our system with several
  baselines and variations.  These experiments demonstrated the
  benefits of our approach in the context of an idealized oracle
  classifier.  We also analyzed where and why our approach fails.  

\item We contrasted our approach with those of several other research
  groups working on incorporating context into object detection, and
  on active object localization.
\end{itemize}
The work described in this paper is an early step in our broader
research goal: to develop a system that integrates cognitive-level
symbolic knowledge with lower-level vision in order to exhibit a deep
understanding of specific visual situations.  This is a long-term and
open-ended project.  In the near-term, we plan to improve our current
system in several ways: 
\begin{itemize}
\item Experimenting with probability models that better fit our target 
  situations, rather than the more limited univariate and multivariate
  normal distributions used here.  We are currently exploring versions
  of kernel density estimation and Gaussian process regression
  algorithms.

\item Replacing the object-proposal scoring oracle with
  category-specific object classifiers, based on convolutional
  network features.  

\item Exploring more sophisticated salience methods to improve
  location priors, while taking into account the tradeoff between the
  usefulness of the location prior and its computational expense.

\item Creating datasets of other visual situations, and evaluating our
  approach on them.  We would like to create a public ``situation
  image dataset repository'' for researchers interested in working on
  recognition of abstract visual situations.  While there are numerous
  action- and event-recognition datasets available, we are not aware of 
  any that are designed specifically to include a very wide variety of
  instances of specific abstract visual situations like those our
  method is aimed at.

\item Comparing our system with related active object localization
  methods on situation-specific image datasets.
\end{itemize}
In the longer term, our goal is to extend Situate to incorporate
important aspects of Hofstadter and Mitchell's Copycat architecture in order 
to give it the ability to quickly and flexibly recognize visual
actions, object groupings, relationships, and to be able to make
analogies (with appropriate conceptual slippages) between a given
image and situation prototypes.  In Copycat, the process of mapping
one (idealized) situation to another was interleaved with the process of building
up a representation of a situation---this interleaving was shown to be
essential to the ability to create appropriate, and even creative
analogies \cite{Mitchell1993}.  Our long-term goal is to build Situate
into a system that bridges the levels of symbolic knowledge and
low-level perception in order to more deeply understand visual
situations---a core component of general intelligence.

\section*{Acknowledgments} 
We are grateful to Jose Escario, Garrett Kenyon, Will Landecker, Sheng
Lundquist, Clinton Olson, Efsun Sarioglu, Kendall Stewart, Mick
Thomure, and Jordan Witte for discussions and assistance concerning
this project.  Many thanks to Li-Yun Wang for running the experiments
with Randomized Prim's algorithm.  This material is based upon work
supported by the National Science Foundation under Grant Number
IIS-1423651.  Any opinions, findings, and conclusions or
recommendations expressed in this material are those of the authors
and do not necessarily reflect the views of the National Science
Foundation.

\bibliographystyle{IEEEtran}
\bibliography{IEEEabrv,SituatePaperArxiv}
 
%%%%%%%%%%%%%%%%%%%%%%%%%%%%%%%%%%%%%%%%%%%%%%%%%%%%%%%%%%%%%%%%%%%%%%%%%%%%%%%%%%%%%%%%%%%%%%%%%%% 

\end{document}